\documentclass[runningheads,a4paper]{llncs}

\usepackage{color}
\usepackage{amssymb}
\usepackage{amsmath}
\usepackage{mathrsfs}
\usepackage{graphicx}
\usepackage{url}
\usepackage{bm}
\usepackage{subfigure}
\usepackage{graphicx}
\usepackage{algorithmic}
\usepackage{algorithm}
\usepackage{siunitx}
\graphicspath{{./pdf/}{./jpg/}{./png/}}
\DeclareGraphicsExtensions{.pdf,.jpg,.png}

\newcommand{\argmax}{\operatornamewithlimits{argmax}}
\newcommand{\argmin}{\operatornamewithlimits{argmin}}

\begin{document}

\mainmatter

\title{Joint inference on structural and diffusion MRI for sequence-adaptive Bayesian segmentation of thalamic nuclei with probabilistic atlases}

\titlerunning{Joint inference on structural and diffusion MRI}
\toctitle{Joint inference on structural and diffusion MRI for sequence-adaptive Bayesian segmentation of thalamic nuclei with probabilistic atlases}

\author{Juan Eugenio Iglesias\inst{1,2,3}, Koen Van Leemput\inst{2,4}, Polina Golland\inst{3}, Anastasia Yendiki\inst{2}}
\authorrunning{Iglesias, Van Leemput, Golland, Yendiki}

\institute{Centre for Medical Image Computing, Department of Medical Physics and Biomedical Engineering, University College London, United Kingdom \\
\and Athinoula A. Martinos Center for Biomedical Imaging, Massachusetts General Hospital and Harvard Medical School, USA \\
\and Computer Science and Artificial Intelligence Laboratory, MIT, USA \\
\and Department of Health Technology, Technical University of Denmark \\}
\tocauthor{Iglesias, Van Leemput, Golland, Yendiki}

\maketitle

\begin{abstract}
Segmentation of structural and diffusion MRI (sMRI/dMRI) is usually performed independently in neuroimaging pipelines. However, some brain structures (e.g., globus pallidus, thalamus and its nuclei) can be extracted more accurately by fusing the two modalities. Following the framework of Bayesian segmentation with probabilistic atlases and unsupervised appearance modeling, we present here a novel algorithm to jointly segment multi-modal sMRI/dMRI data.  We propose a hierarchical likelihood term for the dMRI defined on the unit ball, which combines the Beta and Dimroth-Scheidegger-Watson distributions to model the  data at each voxel.  This term is integrated with a mixture of Gaussians for the sMRI data, such that the resulting joint unsupervised likelihood enables the analysis of multi-modal scans acquired with any type of MRI contrast, b-values, or number of directions, which enables wide applicability.  We also propose an inference algorithm to estimate the maximum-a-posteriori model parameters from input images, and to compute the most likely segmentation.  Using a recently published atlas derived from histology, we apply our method to thalamic nuclei segmentation on two datasets: HCP (state of the art) and ADNI (legacy) -- producing lower sample sizes than Bayesian segmentation with sMRI alone.
\end{abstract}

\section{Introduction}
\label{sec:intro}
Automated segmentation of MRI scans is a prerequisite for most human neuroimaging studies. Most of the algorithms commonly used for this task rely solely on structural MRI (sMRI) scans, and belong to one of three categories: Bayesian segmentation with a probabilistic atlas (e.g.,~\cite{ashburner2005unified,fischl2002whole}); multi-atlas segmentation~\cite{iglesias2015multi}; and, more recently, convolutional neural networks (e.g.,~\cite{roy2018quicknat}). Typically, these techniques segment the brain into tissue types (i.e., gray matter, white matter, and cerebrospinal fluid), or into finer anatomical structures (e.g., hippocampus, ventricle). Bayesian methods drive the primary segmentation modules of the most widespread neuroimaging packages, like FreeSurfer~\cite{fischl2002whole}, FSL~\cite{patenaude2011bayesian}, or SPM~\cite{ashburner2005unified}. 

\begin{figure}[t!]
\centering
\includegraphics[width=1.0\textwidth]{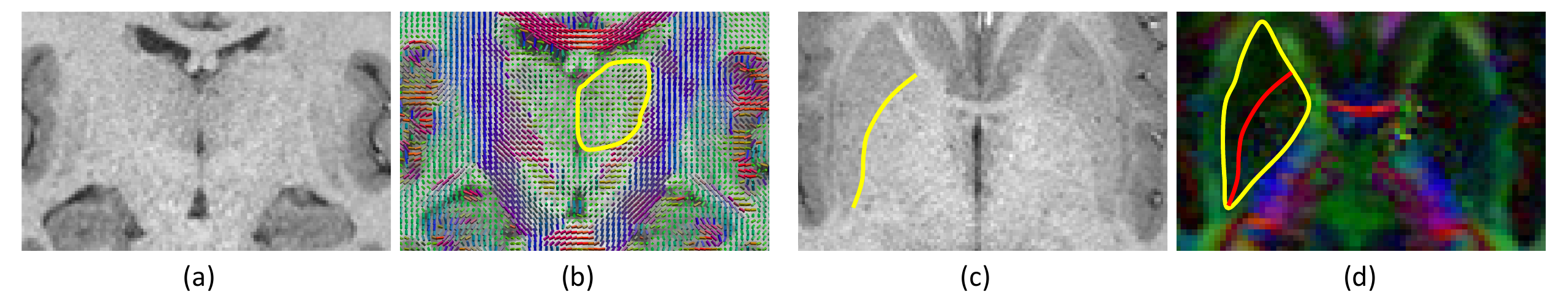}
\caption{(a)~Coronal plane across the thalami of a T1 scan from the HCP. (b)~Corresponding map of  principal diffusion directions, and manual delineation of the left thalamus. (c)~Axial plane of the T1 scan across the basal ganglia, and manual delineation of the boundary between the putamen and globus pallidus.  (d)~Corresponding principal diffusion directions (weighted by FA), with the joint boundary of the putamen and pallidum (in yellow, visible in this map) and the contour from the T1 (in red).}
\label{fig:Motivation}
\end{figure}

The aforementioned approaches rely mostly on T1 contrast to distinguish between gray and white matter. However, some boundaries between structures are nearly invisible in T1  (and other structural MR contrasts) due to insufficient difference in proton density and relaxation times. This is exacerbated by lower contrast-to-noise ratio in deep-brain structures, due to greater distance from the head coil. Two examples from the state-of-the-art Human Connectome Project (HCP) dataset~\cite{van2013wu} are shown in Fig.~\ref{fig:Motivation}. In the first example, the lateral boundary of the thalamus appears very faint (Fig.~\ref{fig:Motivation}a). In the second, the lateral boundary of the globus pallidus is visible thanks to the contrast with the neighboring putamen, but the medial boundary is not (Fig.~\ref{fig:Motivation}c).

These issues create a need for fusing data from several MR modalities to better delineate structure boundaries. A natural complement to sMRI is diffusion MRI (dMRI), which  may help discriminate between certain tissue  types, despite its lower resolution. For example, in Fig.~\ref{fig:Motivation}b,  the lateral boundary of the thalamus is clearly discernible  in  the principal diffusion direction map obtained from dMRI. The diffusion data also complement the T1 scan in the pallidum, which can be delineated by combining contours obtained from the two modalities (medial from dMRI, lateral from sMRI, see Fig.~\ref{fig:Motivation}d). 

Most prior work on segmentation of dMRI focuses on delineating white matter structures, using tractography~\cite{odonnell2007automatic,wassermann2010unsupervised,yendiki2011automated} or volumetric segmentation~\cite{awate2007fuzzy,hagler2009automated}. Tractography has also been used to subdivide subcortical structures (e.g., thalamus~\cite{behrens2003non}, amygdala~\cite{saygin2011connectivity}) based on long-range connections.
Surprisingly, the literature on \emph{joint} modeling of multimodal sMRI/dMRI  is sparse. When sMRI and dMRI are used by the same tool, this is most often done serially, e.g., a segmentation derived from sMRI  is used to analyze the dMRI (e.g., to derive priors for Bayesian tractography~\cite{yendiki2011automated}). To the best of our knowledge, the only works analyzing sMRI and dMRI simultaneously have been on thalamic nuclei segmentation with random forests~\cite{stough2014automatic,glaister2016thalamus}. The main concern with such discriminative techniques is their generalization ability to other datasets, which is limited by differences in MRI acquisition. For sMRI segmentation, this problem can be ameliorated with data augmentation and pretraining~\cite{roy2018quicknat}. However, this is harder to do in dMRI, where acquisition protocols are much less standardized.

The ability to generalize across datasets is critical when software is released publicly and few assumptions can be made on the acquisition. In such scenarios, Bayesian segmentation methods that automatically estimate appearance models from input images remain very popular, as they are agnostic to the MRI contrast of the input scan, and thus robust to acquisition differences. These methods are used for tissue segmentation by major neuroimaging packages (e.g., Unified Segmentation~\cite{ashburner2005unified} in SPM, and FAST~\cite{zhang2001segmentation} in FSL). However, they can be inaccurate when segmenting structures with poor sMRI contrast (see Fig.~\ref{fig:Motivation}).  

Here we propose a sequence-adaptive Bayesian algorithm that uses a probabilistic atlas to segment sMRI and dMRI data \emph{simultaneously}\footnote{Henceforth, we use   ``Bayesian segmentation'' to refer to this specific family  of Bayesian methods, using probabilistic atlases and unsupervised appearance modeling.}. This is achieved via a novel dMRI likelihood term, which relies on a hierarchical model for the fractional anisogropy (FA) and principal diffusion orientation. Combined with a Gaussian likelihood for sMRI, this  model of image intensities is flexible enough to produce accurate segmentations, while keeping  dimensionality low. We also propose a novel inference algorithm to automatically segment scans by fitting the model to multi-modal sMRI/dMRI data. Thanks to unsupervised intensity modeling, applicability across a wide range of acquisition protocols is achieved, which is demonstrated experimentally on two considerably different datasets.

\section{Methods}
\label{sec:methods}

\subsection{Forward probabilistic model}

\begin{figure}[t!]
\centering
\includegraphics[width=0.66\textwidth]{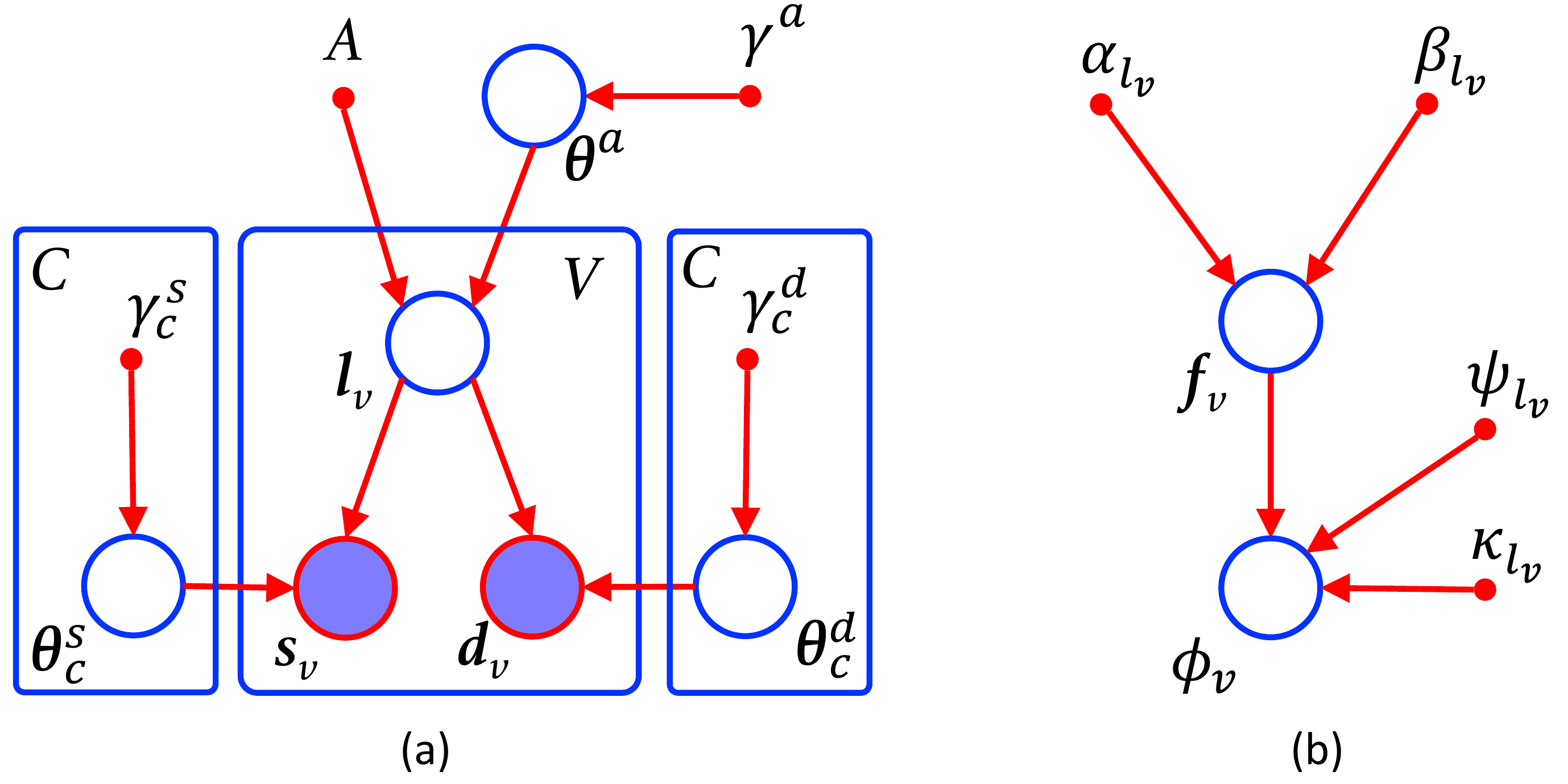}
\caption{(a)~Graphical model of proposed framework. (b)~Hierarchical model of dMRI likelihood. Circles represent random variables  (open if hidden, shaded if observed). Smaller solid circles are deterministic parameters. Plates indicate replication.}
\label{fig:graphicalModel}
\end{figure}

The graphical model of our framework is shown in Fig.~\ref{fig:graphicalModel}a. The observed variables are a bias field corrected (possibly multispectral) sMRI scan  $\bm{S}=[\bm{s}_1,\ldots,\bm{s}_V]$ defined on $V$ voxels, a dMRI scan $\bm{D}=[\bm{d}_1,\ldots,\bm{d}_V]$ defined at the same voxel coordinates (which might require resampling), and a probabilistic atlas $\bm{A}$, which provides the probabilities of observing one of $C$ neuroanatomical classes at every location across a reference spatial coordinate system. The model is governed by three sets of deterministic hyperparameters specified by the user: $\bm{\gamma}^a$, $\bm{\gamma}^s$ and $\bm{\gamma}^d$. 

At the top of the generative model we find the atlas $\bm{A}$, along with a set of related parameters $\bm{\theta}^a$ that deform this atlas into the space of the MRI data. These parameters are a sample of a distribution that  regularizes the deformation field by penalizing, e.g., its bending energy. The strength of the regularization is controlled by the set of hyperparameterst $\bm{\gamma}^a$.

Given the deformed atlas, a labeling (segmentation) $\bm{L}=[l_1,\ldots,l_V]$, with $l_v \in \{1,\ldots,C\}$, is obtained by independently sampling the categorical distribution defined by the deformed atlas at each voxel location. Given  $\bm{L}$, the observed sMRI and dMRI data are assumed to be conditionally independent from each other and across voxels. The sMRI data $\bm{s}_v$ at $v$ follows a distribution (typically a Gaussian) whose parameters $\bm{\theta}^s_c$ depend on the corresponding label $c=l_v$. Any prior knowledge on these parameters is encoded in their priors, which are governed by hyperparameter vectors $\{\bm{\gamma}^s_c\}$. Similarly, $\bm{d}_v$ is also assumed to be a mixture conditioned on the segmentation, described by parameters $\{\bm{\theta}^d_c\}$ and hyperparameters $\{\bm{\gamma}^d_c\}$, which yields a symmetric likelihood model (Fig.~\ref{fig:graphicalModel}a).
The joint probability density function (PDF) of the model is therefore:
\begin{align}
&p(\bm{S}, \bm{D},\bm{L}, \bm{\theta}^a,\bm{\theta}^s,\bm{\theta}^d | \bm{A},\bm{\gamma}^a,\bm{\gamma}^s,\bm{\gamma}^d) \nonumber \\
& = p(\bm{S} | \bm{L}, \bm{\theta}^s) 
  p(\bm{D} | \bm{L}, \bm{\theta}^d) 
  p(\bm{L} | \bm{A}, \bm{\theta}^a)
  p(\bm{\theta}^s | \bm{\gamma}^s) p(\bm{\theta}^d | \bm{\gamma}^d) p(\bm{\theta}^a | \bm{\gamma}^a) \nonumber \\
& =  \left( \prod_{v=1}^V p(\bm{s}_v | \bm{\theta}^s_{l_v})   p(\bm{d}_v | \bm{\theta}^d_{l_v}) p_v(l_v | \bm{A},\bm{\theta}^a) \right)
\left( \prod_{c=1}^C   p(\bm{\theta}^s_c | \bm{\gamma}^s_c ) p(\bm{\theta}^d_c | \bm{\gamma}^d_c ) \right)
p(\bm{\theta}^a | \bm{\gamma}^a), \label{eq:jointPDF}
\end{align}
where $\bm{\theta}^s=\{\bm{\theta}^s_c\}_{c=1,\ldots,C}$, $\bm{\gamma}^s=\{\bm{\gamma}^s_c\}_{c=1,\ldots,C}$, $\bm{\theta}^d=\{\bm{\theta}^d_c\}_{c=1,\ldots,C}$,  $\bm{\gamma}^d=\{\bm{\gamma}^d_c\}_{c=1,\ldots,C}$.

\subsection{Model instantiation}

\subsubsection{Probabilistic atlas:}
We follow the framework of the thalamic atlas~\cite{iglesias2018probabilistic} that we use in the experiments in Section~\ref{sec:ExpAndRes}, in which the atlas is encoded as a tetrahedral mesh. Deforming the mesh is penalized by a regularizer $R$, weighted by the mesh stiffness $\lambda$. The prior is given by (see further details in~\cite{van2009encoding}):
\begin{equation}
p(\bm{\theta}^a | \bm{\gamma}^a) \propto \exp [- \lambda R(\bm{\theta}^a)],
 \quad \text{and} \quad
l_v \sim \text{Cat}[\bm{A}_v(\bm{\theta}^a)], \label{eq:atlasGeneric}
\end{equation}
where $\bm{A}_v(\bm{\theta}^a) = [A_{v1},\ldots,A_{vc}]^t$ is simply the vector of $C$ label probabilities provided by the atlas at voxel $v$ when deformed with parameters $\bm{\theta}^a$;  $\text{Cat}[\cdot]$ is the categorical distribution; and hyperparameters $\bm{\gamma}^a$ comprise just $\lambda$, i.e., $\bm{\gamma}^a = \{\lambda\}$.

\subsubsection{Likelihood of sMRI:}
In order to model the sMRI intensities given the segmentation $\bm{L}$, we follow the Bayesian brain MR segmentation literature (e.g.,~\cite{ashburner2005unified,zhang2001segmentation,iglesias2018probabilistic}) and use Gaussian intensity distributions, such that $\bm{\theta}^s_c=\{\bm{\mu}_c,\bm{\Sigma}_c$\}, i.e., the mean and covariance of the intensities of class $c$. We place a Normal Inverse Wishart (NIW) distribution on these parameters (i.e., the conjugate prior), with zero degrees of freedom for the covariance, as we found it difficult in practice to inform  such parameter a priori. Therefore, we have: $\bm{\gamma}^s_c =\{\bm{M}_c,n_c\}$, where $\bm{M}_c$ is the hypermean and $n_c$ is the scale. The sMRI likelihood is thus:
\begin{align}
\bm{\mu}_c, \bm{\Sigma}_c \enspace | \enspace \bm{M}_c, n_c  &\sim  \text{NIW}(\bm{M}_c, n_c, 0, 0\bm{I}), \nonumber \\
\bm{s}_v  \enspace | \enspace l_v, \{\bm{\mu}_c\}, \{\bm{\Sigma}_c\} &\sim \mathcal{N}(\bm{\mu}_{l_v}, \bm{\Sigma}_{l_v}), \label{eq:lhoodGaussian}
\end{align}  
where $\mathcal{N}(\cdot,\cdot)$ is the Gaussian distribution and $\bm{I}$ is the identity matrix.

\subsubsection{Likelihood of dMRI:}
We have two requirements for the likelihood function of the dMRI: low demands on gradient directions and b-values to accommodate legacy data; and low number of parameters to facilitate unsupervised clustering (yet sufficient to separate the classes). To satisfy the first requirement, we adopt the diffusion tensor imaging (DTI) model, which can be fit from virtually all available dMRI data. Rather than modeling the tensors directly (e.g., with a Wishart distribution, which we found in pilot experiments to fade too quickly from its mode),  we use a hierarchical model (Fig.~\ref{fig:graphicalModel}b) that only considers the FA $f_v$ and the principal eigenvector $\bm{\phi}_{v}$ at each voxel, i.e., $\bm{d}_v = \{f_v,\bm{\phi}_{v}\}$. 

At the first level, we model the FA conditioned on the class, with Beta  distributions parameterized by $\{\alpha_c,\beta_c\}$. We chose the Beta  because it can model location and dispersion of signals defined on the [0,1] interval with two parameters. At the second level, we model the principal eigenvector with the Dimroth-Scheidegger-Watson (DSW) distribution, which is axial (i.e., antipodally symmetric), accommodating the directional invariance of dMRI~\cite{zhang2012noddi}. This distribution is also rotationally symmetric around a mean direction $\bm{\psi}$ and its opposite $-\bm{\psi}$ ($\|\bm{\psi}\|=1$), with a dispersion around the mean parameterized by a concentration $\kappa$. It has fewer parameters than other axial distributions, such as the (non rotationally symmetric) Bingham. Its PDF is given by~\cite{mardia2009directional}: 
\begin{equation}
f(\bm{\phi} ; \bm{\psi}, \kappa) = \left[Z\left(\kappa\right)\right]^{-1} \exp\left[\kappa (\bm{\psi}^t \bm{\phi})^2\right],
\label{eq:DSW}
\end{equation} 
with domain $\|\bm{\phi}\|=1$, and where the partition function is the Kummer function in 3D~\cite{mardia2009directional}: $Z{\kappa}=\int_0^1 \exp(\kappa t^2) dt$. We further assume that the concentration is modulated (multiplied) by the FA. This is a simple yet effective way of modeling the higher directional dispersion in voxels with low FA (e.g., in areas of unrestricted diffusion or fiber crossings), without having to resort to mixtures or additional parameters.  The overall model for the dMRI likelihood is thus:
\begin{align}
f_v \enspace  | \enspace  l_v, \{\alpha_c\}, \{\beta_c\} &\sim \text{Beta}(\alpha_{l_v},\beta_{l_v}), \nonumber \\
\bm{\phi}_v \enspace | \enspace l_v, f_v, \{\bm{\psi}_c\}, \{\kappa_c\} &\sim \text{DSW}(\bm{\psi}_{l_v},f_v\kappa_{l_v})  \label{eq:lhoodDiff}, 
\end{align}
and the set of parameters is thus: $\bm{\theta}_c^d = \{\alpha_c,\beta_c,\bm{\psi}_c,\kappa_c\}$, with $\|\bm{\psi}_c\|=1, \forall c$.   We decided not to inform these parameters, such that $p(\bm{\theta}_c^d)\propto 1$, and $\bm{\gamma_c^d} = \emptyset$. We note that this likelihood model defines a PDF on the unit ball for vector $f_v \bm{\phi}_v$.

\subsection{Segmentation as Bayesian inference}

Within our joint generative model of sMRI and dMRI, we pose segmentation as an optimization problem, seeking to maximize the posterior probability of the labeling, given the known hyperparameters and observed input data:
\begin{align}
\argmax_{\bm{L}} \int \int \int & p(\bm{L} | \bm{\theta}^a, \bm{\theta}^s,\bm{\theta}^d , \bm{S}, \bm{D},\bm{A})  p(\bm{\theta}^a,\bm{\theta}^s,\bm{\theta}^d | \bm{S}, \bm{D},\bm{A},\bm{\gamma}^a,\bm{\gamma}^s) d\bm{\theta}^a d\bm{\theta}^s d\bm{\theta}^d, \nonumber \\
 \approx \argmax_{\bm{L}} \enspace & p(\bm{L} | \hat{\bm{\theta}}^a, \hat{\bm{\theta}}^s,\hat{\bm{\theta}}^d , \bm{S}, \bm{D},\bm{A}), \label{eq:approxInference}
\end{align}
where we have made the standard approximation that the posterior distribution of the parameters is heavily peaked around point estimates $\hat{\bm{\theta}}^a$, $\hat{\bm{\theta}}^s$, $\hat{\bm{\theta}}^d$  given by:
\begin{equation}
\{\hat{\bm{\theta}}^a,\hat{\bm{\theta}}^s,\hat{\bm{\theta}}^d\} = \argmax_{\{{\bm{\theta}}^a,{\bm{\theta}}^s,{\bm{\theta}}^d\}} p(\bm{\theta}^a,\bm{\theta}^s,\bm{\theta}^d | \bm{S}, \bm{D},\bm{A},\bm{\gamma}^a,\bm{\gamma}^s). \label{eq:pointEstimates}
\end{equation}
Therefore, we segment a scan by first estimating the parameters with Eq.~\ref{eq:pointEstimates}, and then obtaining the (approximate) most likely labeling with Eq.~\ref{eq:approxInference}.  

Applying Bayes rule to Eq.~\ref{eq:pointEstimates},  marginalizing over the hidden segmentation $\bm{L}$, and considering the structure of the model and our design choices, we obtain:
$$
\{\hat{\bm{\theta}}^a,\hat{\bm{\theta}}^s,\hat{\bm{\theta}}^d\} = \argmax_{\{{\bm{\theta}}^a,{\bm{\theta}}^s,{\bm{\theta}}^d\}}
p(\bm{\theta}^a | \bm{\gamma}^a) p(\bm{\theta}^s|\bm{\gamma}^s) 
\sum_{\bm{L}} p(\bm{S} | \bm{L},\bm{\theta}^s) p(\bm{D} | \bm{L},\bm{\theta}^d) 
p(\bm{L} | \bm{\theta}^a,\bm{A}). 
$$
Expanding and taking logarithm, we obtain the following objective function: 
\begin{align}
O&({{\bm{\theta}}^a,\{{\bm{\mu}_c},{\bm{\Sigma}_c},{\alpha}_c,{\beta}_c,{\bm{\psi}}_c,{\kappa}_c\}}) =  \log p(\bm{\theta}^a | \bm{\gamma}^a) + \sum_{c=1}^C \log p(\bm{\mu}_c,\bm{\Sigma}_c | \bm{M}_c, n_c) \nonumber \\
&+ \sum_{v=1}^V \log \left[ \sum_{c=1}^C p(\bm{s}_v | \bm{\mu}_c,\bm{\Sigma}_c) p(f_v | \alpha_c, \beta_c) p(\bm{\phi}_v | f_v, \bm{\psi}_c, \kappa_c) p( l_v = c | \bm{A},\bm{\theta}^a) \right]. \label{eq:objective}
\end{align}
We maximize Eq.~\ref{eq:objective} with a Generalized Expectation Maximization (GEM) algorithm~\cite{dempster1977maximum},
iterating between expectation (E) and maximization (M) steps:

\subsubsection{E step: }
In the E step, we use Jensen's inequality to build a lower bound for the objective function, which touches it at the current value of the parameters:
\begin{align}
O  \geq Q & =  \sum_{v=1}^V \sum_{c=1}^C w_{vc} \log \left[  p(\bm{s}_v | \bm{\mu}_c,\bm{\Sigma}_c) p(f_v | \alpha_c, \beta_c) p(\bm{\phi}_v | f_v, \bm{\psi}_c, \kappa_c) p( l_v = c | \bm{A},\bm{\theta}^a) \right] \nonumber \\
&+ \log p(\bm{\theta}^a | \bm{\gamma}^a) + \sum_{c=1}^C \log p(\bm{\mu}_c,\bm{\Sigma}_c | \bm{M}_c, n_c) -\sum_{v=1}^V \sum_{c=1}^C w_{vc} \log w_{vc}, \label{eq:boundEstep}
\end{align}
where $\{w_{vc}\}$ a soft segmentation according to the current parameter estimates:
\begin{align}
w'_{vc} & =  p(\bm{s}_v | \bm{\mu}_c,\bm{\Sigma}_c) p(f_v | \alpha_c, \beta_c) p(\bm{\phi}_v | f_v, \bm{\psi}_c, \kappa_c) p( l_v = c | \bm{A},\bm{\theta}^a) \nonumber \\
  = & |\bm{\Sigma}_c|^{-1/2} \exp[-\frac{1}{2}(\bm{s}_v - \bm{\mu}_c)^t  \bm{\Sigma}_c^{-1} (\bm{s}_v - \bm{\mu}_c) ]  
f_v^{\alpha_c-1} (1-f_v)^{\beta_c-1} [B(\alpha_c,\beta_c)]^{-1} \nonumber \\
\times & [Z(f_v\kappa_c)]^{-1} \exp[f_v\kappa_c (\bm{\psi}_c^t \bm{\phi}_v)^2] A_{vc}(\bm{\theta}^a), \enspace \text{and} \enspace w_{vc} =  w'_{vc} /  \sum_{c'=1}^C w'_{vc'}, \label{eq:E-step}
\end{align}
where $B$ is the Beta function.

\subsubsection{M step: }
In the generalized M step, we seek to improve the lower bound Q in Eq.~\ref{eq:boundEstep}. While optimizing the bound with respect to all parameters simultaneously is difficult, optimizing different subsets each time (coordinate ascent) is feasible.

\smallskip
\noindent\textit{Optimizing $\bm{\theta}^a$: } 
Fixing all other parameters and switching signs, we obtain:
\begin{equation}
\argmin_{\bm{\theta}^a} \enspace  \sum_{v=1}^V \sum_{c=1}^C w_{vc} \log \left[ w_{vc} / A_{vc}(\bm{\theta}^a)\right] + \lambda R(\bm{\theta}^a).  \label{eq:MstepDeformation}  
\end{equation}
This is a  registration problem combining the regularizer $R$ with a data term: the Kullback--Leibler (KL) divergence between the deformed atlas and the current soft segmentation. We solve it numerically with the conjugate gradient method.

\smallskip 
\noindent\textit{Optimizing $\{\bm{\mu}_c,\bm{\Sigma}_c\}$: } Setting derivatives to zero yields a closed-form solution:
\begin{align}
\bm{\mu}_c & = \frac{n_c \bm{M}_c + \sum_{v=1}^{V} w_{vc} \bm{s}_v}{n_c+ \sum_{v=1}^{V} w_{vc} }, \label{eq:MstepMean} \\
\bm{\Sigma}_c & = \frac{n_c (\bm{\mu}_c - \bm{M}_c ) (\bm{\mu}_c - \bm{M}_c )^t+ \sum_{v=1}^{V} w_{vc} (\bm{s}_v-\bm{\mu}_c) (\bm{s}_v-\bm{\mu}_c)^t}{1+ \sum_{v=1}^{V} w_{vc} } \label{eq:MstepSigma}.
\end{align}

\noindent\textit{Optimizing $\{\alpha_c,\beta_c\}$: }  Substituting the expression of the Beta distribution  into Eq.~\ref{eq:boundEstep}, the problem decouples across classes:
\begin{equation}
\argmax_{\alpha_c,\beta_c}   (\alpha_c-1) \sum_{v=1}^V w_{vc} \log f_v + (\beta_c-1) \sum_{v=1}^V w_{vc} \log (1-f_v) - \log B(\alpha_c,\beta_c)  \sum_{v=1}^V w_{vc}. \label{eq:MstepBeta}
\end{equation}
This is a simple 2D optimization problem, which we solve with conjugate gradient. In the first iteration, we use the method of moments for initialization.

\smallskip
\noindent\textit{Optimizing $\{\bm{\psi}_c\}$: } This optimization can also be carried out one $c$ at the time:
$$
\argmax_{\bm{\psi}_c:  \|\bm{\psi}_c\|=1} \sum_v w_{vc} f_v (\bm{\psi}^t_c \bm{\phi}_v)^2 = \argmax_{\bm{\psi}_c:  \|\bm{\psi}_c\|=1}  \bm{\psi}^t_c [ \sum_v w_{vc} f_v  \bm{\phi}_v \bm{\phi}_v^t   ]  \bm{\psi}_c,
$$
with closed-form solution  given by the leading eigenvector of: $\sum_v w_{vc} f_v \bm{\phi}_v \bm{\phi}_v^t$.

\smallskip
\noindent\textit{Optimizing $\{\kappa_c\}$: } This optimization problem also decouples across classes:
\begin{equation}
\argmax_{\kappa_c} \enspace \kappa_c \sum_{v=1}^V w_{vc} f_v (\bm{\psi}^t_c \bm{\phi}_v)^2 - \sum_{v=1}^V w_{vc} \log Z(f_v \kappa_c), \label{eq:MstepKappa}
\end{equation}
which we solve with conjugate gradient, initializing $\kappa_c=10$ in the first iteration.

\subsubsection{Final Segmentation: } It is straightforward to show that the approximate posterior probability of the segmentation from Eq.~\ref{eq:approxInference} factorizes across voxels and is given by 
$
 p(\bm{L} | \hat{\bm{\theta}}^a, \hat{\bm{\theta}}^s,\hat{\bm{\theta}}^d , \bm{S}, \bm{D},\bm{A},\bm{\gamma}^a,\bm{\gamma}^s)  = \prod_{v=1}^V \hat{w}_{v,l_j},
$
where $\hat{w}_{v,l_j}$ is obtained by evaluating Eq.~\ref{eq:E-step} at the optimal parameter values $\hat{\bm{\theta}}^a,\hat{\bm{\theta}}^s,\hat{\bm{\theta}}^d$. Therefore, the optimal segmentation can be computed independently at each location $v$ as: 
\begin{equation}
\hat{l}_v = \argmax_c \enspace \hat{w}_{vc},
\label{eq:hardSeg}
\end{equation}
and the expected value of the volume of class $c$ is given by: $\sum_{v=1}^V  \hat{w}_{vc}$ (in voxels).

\subsubsection{Implementation details: }
Since GEM only requires \emph{improving} the bound at each iteration, 
we follow a schedule in which all the model parameters except for $\bm{\theta}^a$ are updated once in the M step. Since updating $\bm{\theta}^a$ requires solving a more computationally expensive registration problem, we only update $\bm{\theta}^a$ in the M step every five GEM iterations. The method is summarized in Algorithm~\ref{alg:method}.

In practice, we also force some parameters $\{\bm{\theta}^s_c\}$ and $\{\bm{\theta}^d_c\}$ to be shared across classes, for increased robustness of the algorithm. For the sMRI parameters ($\{\mu_c,\sigma^2_c\}$), we follow~\cite{iglesias2018probabilistic} and force parameter sharing across: cortex, hippocampus and amygdala; reticular nucleus and white matter; mediodorsal and pulvinar nuclei;  rest of thalamic nuclei; and contralateral structures. For the FA, parameters $\{\alpha_c,\beta_c\}$ are shared across each of the  six groups of thalamic nuclei in Table~2 of~\cite{iglesias2018probabilistic}, and across contralateral structures. The same grouping -- but without contralateral constraints -- is used for the directional parameters $\{\bm{\psi}_c,\kappa_c\}$.  

\begin{algorithm}[!t]
 \caption{Bayesian segmentation with sMRI and dMRI}
 \label{alg:method}
 \begin{algorithmic}
 \REQUIRE  $\bm{A}, \bm{S}, \bm{D}, \bm{\gamma}^a, \bm{\gamma}^s, \{\bm{M}_c,n_c\}$ 
 \ENSURE $\hat{\bm{\theta}}^a,\{\hat{\bm{\mu}_c},\hat{\bm{\Sigma}_c},\hat{\alpha}_c,\hat{\beta}_c,\hat{\bm{\psi}}_c,\hat{\kappa}_c\}$
 \STATE Initialize $\bm{\theta}^a$, with affine registration and mutual information
 \STATE Initialize $w_{vc} \leftarrow p(l_v = c | \bm{A}, \bm{\theta}^a), \forall v,c$
 \STATE Initialize $\kappa_c=10, \forall c$
 \STATE Initialize $\alpha_c, \beta_c$ with method of moments, $\forall c$
 \STATE $\text{it} \leftarrow 0$
 \WHILE{${\bm{\theta}}^a,\{{\bm{\mu}_c},{\bm{\Sigma}_c},{\alpha}_c,{\beta}_c,{\bm{\psi}}_c,{\kappa}_c\}$ change AND $\text{it} \leq \text{it}_\text{max}$}
 \STATE $\text{it} \leftarrow \text{it} + 1$
  \FOR{$c=1,\ldots,C$}
 \STATE Update $\bm{\mu}_c$, $\bm{\Sigma}_c$  with Eqns.~\ref{eq:MstepMean} and~\ref{eq:MstepSigma}
 \STATE Update $\alpha_c, \beta_c$ by numerically optimizing Eq.~\ref{eq:MstepBeta} with conjugate gradient 
 \STATE Update $\bm{\psi}_c \leftarrow \text{leading eigenvector of:} \enspace \sum_v w_{vc} f_v \bm{\phi}_v \bm{\phi}_v^t$
 \STATE Update $\kappa_c$ by numerically optimizing Eq.~\ref{eq:MstepKappa} with conjugate gradient 
 \ENDFOR
 \IF{mod(its,5)=0}
 \STATE Update $\bm{\theta}^a$ by numerically optimizing Eq.~\ref{eq:MstepDeformation} with conjugate gradient 
 \ENDIF
 \STATE Update $w_{vc}$ with Eq.~\ref{eq:E-step}, $\forall v,c$
 \ENDWHILE
 \STATE Compute final segmentation with Eq.~\ref{eq:hardSeg}.
  \end{algorithmic}
 \end{algorithm}

\section{Experiments and results} 
\label{sec:ExpAndRes}

\subsection{Data}

We evaluate our method with a recent probabilistic atlas of 25 thalamic nuclei and surrounding regions derived from histology~\cite{iglesias2018probabilistic}. The thalamus is an excellent target region, due to its faint lateral boundaries in sMRI (as explained in Section~\ref{sec:intro}), and its set of nuclei with different connectivity. We use two considerably different datasets in evaluation: HCP (state of the art) and ADNI (legacy). 

\smallskip
\noindent\textit{HCP: } 
Isotropic T1 and dMRI scans from 100 healthy subjects (age 29.1$\pm$3.3, 44 males), at 0.7 mm  (T1) and 1.25 mm resolution (dMRI). We fit the DTI model to the $b$=1000 s/mm$^2$ shell (180 directions) and 12 scans with $b$=0 (details in~\cite{sotiropoulos2013advances}). 

\smallskip
\noindent\textit{ADNI: } 
T1 and dMRI scans from 77 subjects from ADNI2: 39 Alzheimer's disease (AD) and 38 age-matched controls (74.1$\pm$8.1 years; 40 females total). T1 resolution: 1.2$\times$1$\times$1 mm (sagittal); dMRI resolution:1.35$\times$1.35$\times$2.7 mm (axial); 5 volumes with b=0, 41 directions (b=1000 s/mm$^2$; details at \url{adni-info.org}).

\subsection{Experimental setup}

We evaluate three competing methods: {\it (i)}~Segmentation of the whole thalamus with FreeSurfer~\cite{fischl2002whole}; {\it (ii)}~Segmentation of thalamic nuclei using Bayesian segmentation on T1 only~\cite{iglesias2018probabilistic}; and {\it (iii)}~Segmentation of thalamic nuclei with the full model, including dMRI. We compare these approaches in three experiments: {\it (i)}~Qualitative assessment of segmentation and tractography in HCP; {\it (ii)}~Correlation between thalamic and total intracranial volume (ICV) in HCP; and {\it (iii)}~Ability to discriminate AD and control subjects based on volumes in ADNI. The sMRI and dMRI data are resampled to 0.5 mm isotropic in a bounding box around the thalami (DTI is interpolated in a log-euclidean framework~\cite{arsigny2006log}). We set $\lambda=0.05$ as in~\cite{iglesias2018probabilistic}, $M_c$ to the median T1 intensity in class $c$ according to the main FreeSurfer segmentation, and $n_c$ to the volume of the class in mm$^3$.

\subsection{Results}
\label{sec:results}

\begin{figure}[t!]
\centering
\includegraphics[width=0.879\textwidth]{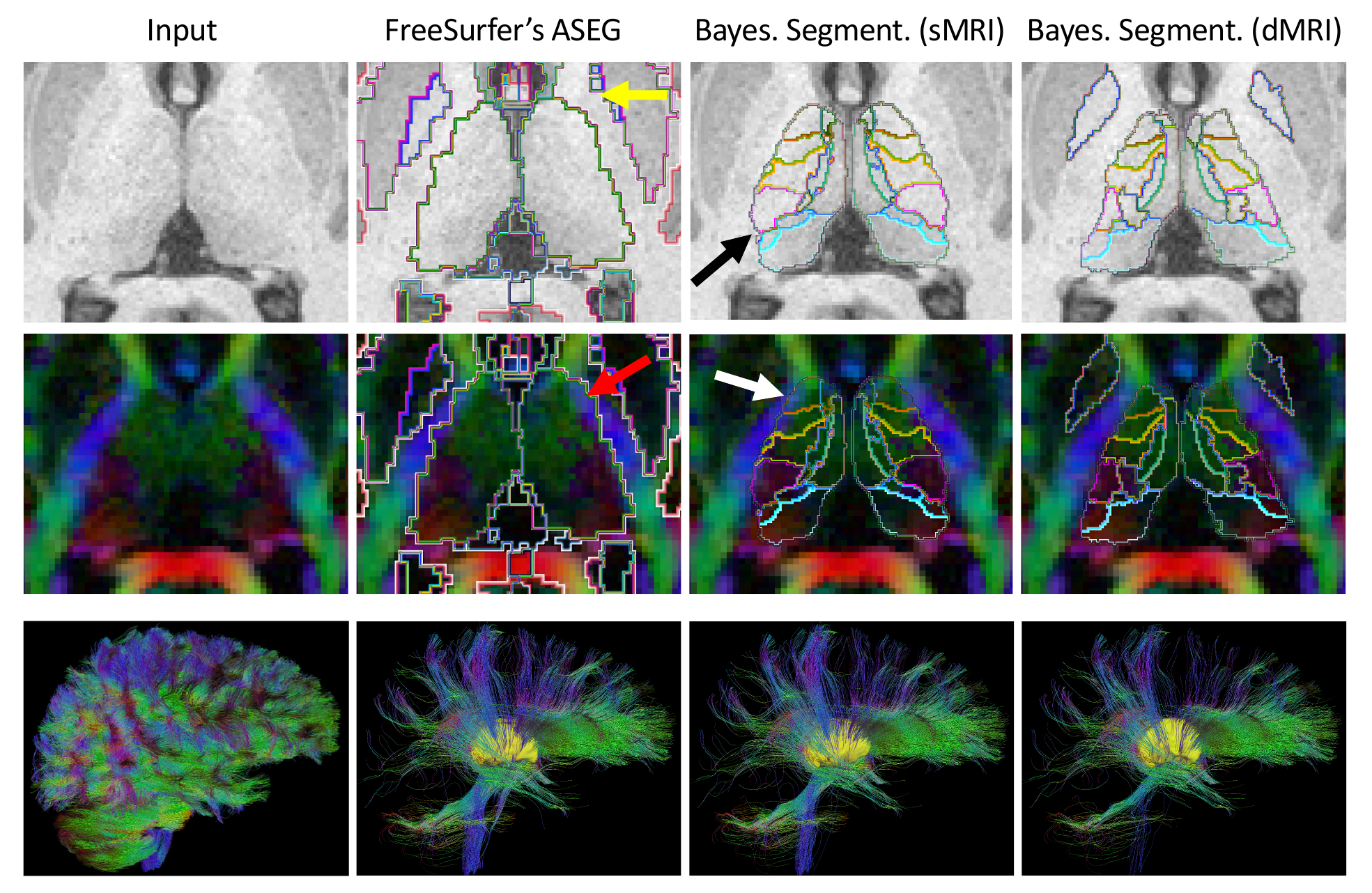}
\caption{Top two rows: axial slice of a T1 scan and principal diffusion directions of an HCP subject, with segmentations from FreeSurfer, Bayesian segmentation (T1 only), and the proposed method (T1+dMRI). Bottom row (left to right): Whole-brain tractography (25,000 tracts); subset of tracts going through the thalami (in yellow) as segmented by: FreeSurfer (2,602 tracts);  Bayesian segmentation of T1 (2,193 tracts); and proposed method (1,676 tracts). See Section~\ref{sec:results} for a description of the arrows.}
\label{fig:qualitative}
\end{figure}

Figure~\ref{fig:qualitative} shows qualitative results on an HCP subject. FreeSurfer almost completely misses the left pallidum (yellow arrow in the figure) and oversegments the thalami. We test the effects of the latter on tractography by reconstructing the full dMRI data with generalized q-sampling~\cite{yeh2010generalized}, performing whole-brain tractography, and isolating the tracts that intersect the whole thalami, as automatically segmented by the three competing methods. The FreeSurfer thalamus yields many false positive tracts, mostly due to overlap with the internal capsule (red arrow). Aggregating the nuclei produced by Bayesian segmentation on the T1 produces a more accurate boundary, but still oversegments the anterior thalamus (white arrow), and undersegments the  pulvinar nucleus (black arrow). Our  multi-modal method yields less false positive tracts, and segments thalamic nuclei that are more homogeneous in terms of diffusion orientation and FA.

\begin{figure}[t!]
\centering
\includegraphics[width=0.879\textwidth]{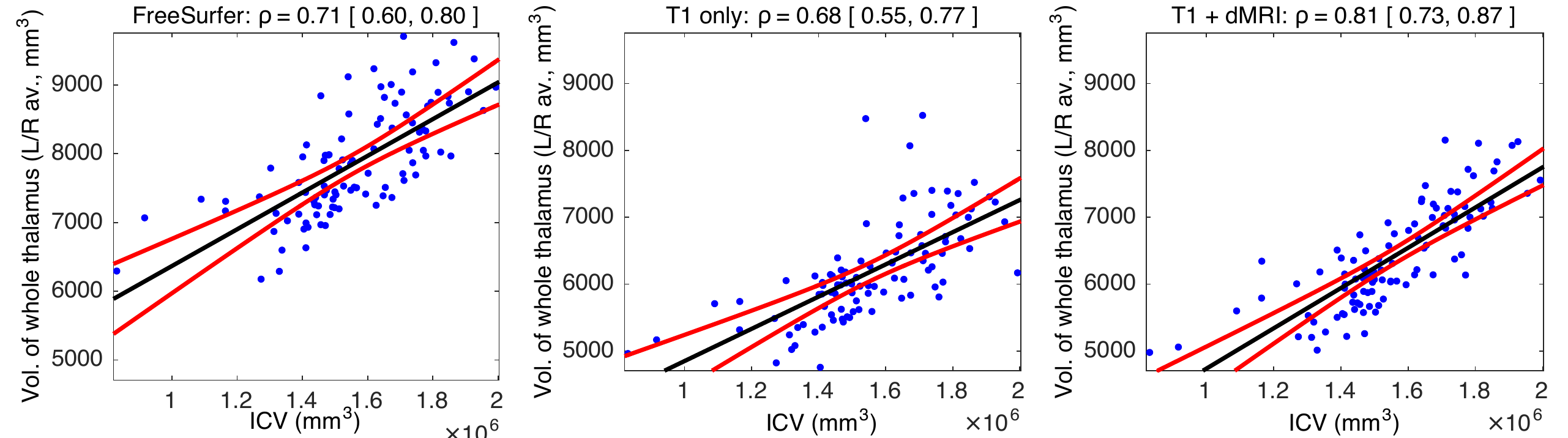}
\caption{Scatter plot for intracranial vs. whole thalamic volumes (left-right averaged), and regression lines (black) with 95\% confidence intervals (in red).}
\label{fig:correlations}
\end{figure}

We also evaluate segmentation performance  quantitatively on HCP,  in an indirect fashion, by computing the correlation of total thalamic volume  obtained by each method (left-right averaged) with the ICV estimated by FreeSurfer; noisy thalamic segmentations are expected to degrade this correlation. Scatter plots and regression lines are shown in Fig.~\ref{fig:correlations}. The FreeSurfer volumes are quite large on average, and their correlation with ICV is $\rho$=0.71. Bayesian segmentation with T1 yields $\rho$=0.68 (not significantly different, with p=0.37 on a two-tailed Steiger test). 
The proposed algorithm produces fewer outliers than the other two, and yields the highest correlation ($\rho$=0.81), significantly higher than those of FreeSurfer (p=0.006) and T1-only segmentation (p=0.001). 

Finally, we evaluate the ability of the segmented volumes to classify the ADNI subjects into AD and controls. We use a simple linear discriminant analysis, 
whose performance is mostly determined by the quality of the volumes. We project the volumes (left-right averaged, corrected for ICV and age) onto the normal to the discriminant hyperplane in a leave-one-out fashion. We use the projections to compute the area under the ROC curve (AUC), accuracy at its elbow, and sample sizes ($\alpha=0.05,\beta=0.2$). Results are shown in Table~\ref{tab:AUROC}. Our method yields a fair improvement compared with T1-only Bayesian segmentation (increase of 7 points in accuracy and AUC, and reduction of 6 samples). Compared with FreeSurfer, our algorithm reduces the sample size by 60\%. 

\begin{table}[t!]
\centering
\caption{AUC, accuracy at elbow, and sample size for the AD classification experiment.}
\begin{tabular}{|c|c|c|c|}
\hline
Method & FreeSurfer (whole)  & Bayes. Seg. T1 & Proposed \\
\hline
AUC  & 60.3\% & 66.5\% & 73.6\% \\
Accuracy at elbow & 61.0\% & 67.5\% & 74.0\%  \\
Sample size & 50 & 26 & 20  \\
\hline
\end{tabular}
\label{tab:AUROC}
\end{table}

\section{Conclusion}
\label{sec:conclusion}

We have presented a Bayesian method for  joint segmentation of sMRI and dMRI, which is robust to changes in acquisition platform and protocol -- as shown with two substantially different datasets. 
Compared with Bayesian segmentation using sMRI alone, our method produces more accurate boundaries for subcortical structures, and yields smaller sample sizes  in an AD classification task. 

Future work at the methodological level will follow five main directions: {\it(i)}~Modeling partial voluming in the dMRI, which may be important for smaller structures; {\it(ii)}~Exploring other axial PDFs, as well as mixtures; {\it(iii)}~Placing a prior on the dMRI likelihood parameters, e.g., to utilize prior knowledge on the FA; {\it(iv)}~Modeling the bias field in the sMRI data, e.g., as in~\cite{ashburner2005unified}; and {\it(v)}~Adding connectivity derived from tractography to the dMRI likelihood, which may be challenging because tractography results depend largely on the MR acquisition.

We also plan to manually trace structures some of the HCP and AD data, with three purposes. First, to include white matter bundles in the atlas, as modeling the whole cerebral white matter with a single Beta-DSW is not realistic (not even within a bounding box). Second, to enable direct evaluation of our  segmentations. And third, to help us explain discrepancies in AD classification accuracy between our results and those presented in~\cite{iglesias2018probabilistic}, which may be due to the their larger dataset, their different ADNI sample, or some other factor.

As high-resolution dMRI becomes more common in neuroimaging, we believe that segmentation techniques that jointly model gray and white matter with sMRI and dMRI -- like the one in this paper -- will become increasingly important.

\medskip
\noindent\textbf{Acknowledgement:} Supported by Horizon 2020 (ERC Starting Grant 677697, Marie Curie grant 765148), Danish Council for Independent Research (DFF-6111-00291), NIH (R21AG050122, P41EB015902), Wistron Corp., SIP, and AWS.

\bibliographystyle{splncs}
\bibliography{bibliography}
\end{document}